# PARSING AS TREE TRAVERSAL

Dale Gerdemann*

Seminar für Sprachwissenschaft, Universität Tübingen[†]

**ABSTRACT**

This paper presents a unified approach to parsing, in which top-down, bottom-up and left-corner parsers are related to preorder, postorder and inorder tree traversals. It is shown that the simplest bottom-up and left-corner parsers are left recursive and must be converted using an extended Greibach normal form. With further partial execution, the bottom-up and left-corner parsers collapse together as in the BUP parser of Matsumoto.

## 1  INTRODUCTION

In this paper, I present a unified approach to parsing, in which top-down, bottom-up and left-corner parsers are related to preorder, postorder and inorder tree traversals. To some extent, this connection is already clear since for each parsing strategy the nodes of the parse tree are constructed according to the corresponding tree traversal. It is somewhat trickier though, to actually use a tree traversal program as a parser since the resulting parser may be left recursive. This left recursion can be eliminated, however, by employing a version of Greibach Normal Form which is extended to handle argument instantiations in definite clause grammars.

The resulting parsers resemble the standard Prolog versions of versions of such parsers. One can then go one step further and partially execute the parser with respect to a particular grammar—as is normally done with definite clause grammars (Pereira & Warren [10]). A surprising result of this partial execution is that the bottom-up and left-corner parsers become identical when they are both partially executed. This may explain why the BUP parser of Matsumoto et al. [6] [7] was referred to as a bottom-up parser even though it clearly follows a left-corner strategy.

## 2  TREE TRAVERSAL PROGRAMS

Following O'Keefe [8], we can implement preorder, postorder and inorder tree traversals as DCGs, which will then be converted directly into top-down bottom-up and left-corner parsers, respectively. The general schema is:

> x_order(Tree) →
> ⟨*x_ordered node labels in Tree*⟩.

Note that in this case, since we are most likely to call `x_order` with the `Tree` variable instantiated, we are using the DCG in generation mode rather than as a parser. When used as a parser

*The research presented in this paper was partially sponsored by Teilprojekt B4 "Constraints on Grammar for Efficient Generation" of the Sonderforschungsbereich 340 of the Deutsche Forschungsgemeinschaft. I would also like to thank Guido Minnen and Dieter Martini for helpful comments. All mistakes are of course my own.

[†]Kl. Wilhelmstr. 113, D-72074 Tübingen, Germany, dg@sfs.nphil.uni-tuebingen.de.

cmp-lg/9407027  29 Jul 94

on the string *S*, the procedure will return all trees whose `x_order` traversal produces *S*. The three instantiations of this procedure are as follows:

```
% preorder traversal
pre(empty) --> [].
pre(node(Mother,Left,Right)) -->
    [Mother],
    pre(Left),
    pre(Right).

% postorder traversal
post(empty) --> [].
post(node(Mother,Left,Right)) -->
    post(Left),
    post(Right),
    [Mother].

% inorder traversal
in(empty) --> [].
in(node(Mother,Left,Right)) -->
    in(Left),
    [Mother],
    in(Right).
```

## 2.1 DIRECT ENCODING OF PARSING STRATEGIES

Analogous to these three traversal programs, there are three parsing stragegies, which differ from the tree traversal programs in only two respects. First, the base case for a parser should be to parse a lexical item rather than to parse an empty string. And second, in the recursive clauses, the mother category fits into the parse tree and is licensed by the auxiliary predicate `rule/3` but it does not figure into the string that is parsed.

As was the case for the three tree traversal programs, the three parsers differ from each other only with respect to the right hand side order. For simplicity, I assume that phrase structure rules are binary branching, though the approach can easily be generalized to non-binary branching.[1]

```
% top-down parser
td(node(PreTerm,lf(Word))) -->
    [Word],
    {word(PreTerm,Word)}.
td(node(Mother,Left,Right)) -->
    {rule(Mother,Left,Right)},
    td(Left),
    td(Right).

% bottom-up parser
bu(node(PreTerm,lf(Word))) -->
    [Word],
    {word(PreTerm,Word)}.
bu(node(Mother,Left,Right)) -->
    bu(Left),
    bu(Right),
    {rule(Mother,Left,Right)}.

% left-corner parser
lc(node(PreTerm,lf(Word))) -->
    [Word],
    {word(PreTerm,Word)}.
lc(node(Mother,Left,Right)) -->
    lc(Left),
    {rule(Mother,Left,Right)},
    lc(Right).
```

As seen here the only difference between the three strategies concerns the choice of when to select a phrase structure rule.[2] Do you start with a rule and then try to satisfy it as in the top-down approach, or do you parse the daughters of a rule first before selecting the rule as in the bottom-up approach, or do you take an intermediate strategy as in the left-corner approach.

---

[1] The only problematic case is for left corner since the corresponding tree traversal *inorder* is normally defined only for binary trees. But inorder is easily extended to non-binary trees as follows: i. visit the left daughter in inorder, ii. visit the mother, iii. visit the rest of the daughters in inorder.

[2] As opposed to, say, a choice of whether to use operations of expanding and matching or operations of shifting and reducing.

# 3 GREIBACH NORMAL FORM PARSERS

While this approach reflects the logic of the top-down, bottom-up and left-corner parsers in a clear way, the resulting programs are not all usable in Prolog since the bottom-up and the left-corner parsers are left-recursive. There exists, however, a general technique for removal of left-recursion, namely, conversion to Greibach normal form. The standard Greibach normal form conversion, however, does not allow for DCG type rules, but we can easily take care of the Prolog arguments by a technique suggested by Problem 3.18 of Pereira & Shieber [9] to produce what I will call *Extended Greibach Normal Form* (EGNF).[3] Pereira & Shieber's idea has been more formally presented in the *Generalized Greibach Normal Form* of Dymetman ([1] [2]), however, the simplicity of the parsers here does not justify the extra complication in Dymetman's procedure. Using this transformation, the bottom-up parser then becomes as follows:[4]

---

[3] EGNF is similar to normal GNF except that the arguments attached to non-terminals must be manipulated so that the original instantiations are preserved. For specific grammars, it is pretty easy to see that such a manipulation is possible. It is much more difficult (and beyond the scope of this paper) to show that there is a general rule for such manipulations.

[4] The Greibach NF conversion introduces one auxiliary predicate, which (following Hopcroft & Ullman [4]) I have called b. Of course, the GNF conversion also does not tell us what to do with the auxiliary procedures in curly brackets. What I've done here is simply to put these auxiliary procedures in the transformed grammar in positions corresponding to where they occurred in the original grammar. It's not clear that one can always find such a "corresponding" position, though in the case of the bottom-up and left-corner parsers such a position is easy to identify.

```
% EGNF bottom-up
bu(node(PreTerm,lf(Word))) -->
    [Word],
    {word(PreTerm,Word)}.
bu(Node) -->
    [Word],
    {word(PreTerm,Word)},
    b(node(PreTerm,lf(Word)),Node).

b(L,node(Mother,L,R)) -->
    bu(R),
    {rule(Mother,L,R)}.
b(L,Node) -->
    bu(R),
    {rule(Mother,L,R)},
    b(node(Mother,L,R),Node).
```

This, however is not very efficient since the two clauses of both bu and b differ only in whether or not there is a final call to b. We can reduce the amount of backtracking by encoding this optionality in the b procedure itself.

```
% Improved EGNF bottom-up
bu(Node) -->
    [Word],
    {word(PreTerm,Word)},
    b(node(PreTerm,lf(Word)),Node).

b(Node,Node) --> [].
b(L,Node) -->
    bu(R),
    {rule(Mother,L,R)},
    b(node(Mother,L,R),Node).
```

By the same EGNF transformation and improvements, the resulting left-corner parser is only minimally different from the bottom-up parser:

```
% Improved EGNF Left-corner
lc(Node) -->
    [Word],
    {word(PreTerm,Word)},
    b(node(PreTerm,lf(Word)),Node).
```

```
b(Node,Node) --> [].
b(L,Node) -->
  {rule(Mother,L,R)},
  lc(R),
  b(node(Mother,L,R),Node).
```

## 4 PARTIAL EXECUTION

The improved EGNF bottom-up and left-corner parsers differ now only in the position of the auxiliary predicate in curly brackets. If this auxiliary predicate is partially executed out with respect to a particular grammar, the two parsers will become identical. For example, if we have a rule of the form:

```
s(tree(s,NP,VP)) -->
   np(NP),
   vp(VP).
```

For either parser, this will result in one b clause of the form:

```
b(np(NP),Node) -->
  lc(vp(VP)),
  b(node(s(tree(s,NP,VP)),
         np(NP),vp(VP)),Node).
```

This is essentially equivalent to the kind of rules produced by Matsumoto et al. ([6] [7]) in their "bottom-up" parser BUP.[5] As seen here, Matsumoto et al were not wrong to call their parser bottom-up, but they could have just as well called it left-corner.

## 5 CONCLUSION

In most standard presentations, simple top-down, bottom-up and left-corner parsers are described in terms of pairs of operations such as expand/match, shift/reduce or sprout/match. But it is entirely unclear what expanding and matching has to do with shifting, reducing or sprouting. By relating parsing to tree traversal, however, it becomes much clearer how these three approaches to parsing relate to each other. This is a natural comparison, since clearly the possible orders in which a tree can be traversed should not differ from the possible orders in which a parse tree can be constructed. What's new in this paper, however, is the idea that such tree traversal programs could be translated into parsers using extended Greibach Normal Form.

Such a unified approach to parsing is mostly useful simply to understand how the different parsers are related. It is surprising to see, for example, that with partial execution, the bottom-up and left-corner parsers become the same. The similarity between bottom-up and left-corner parsing has caused a certain amount of confusion in the literature. For example, the so-called "bottom-up" chart parser presented (among other places) in Gazdar & Mellish [3] in fact uses a left-corner strategy. This was pointed out by Wiren [11] but has not received much attention in the literature. It is hoped that the unified approach to parsing presented here will help to clear up other such confusions.

Finally, one might mention a connection to Government-Binding parsing as presented in Johnson & Stabler [5]. These authors present a generate and test approach, in which X-bar structures are randomly generated and then tested against GB principles. Once the logic of the program is expressed in such a manner, efficiency considerations are used in order to fold the testing procedures into the generation procedure.

---

[5]This rule is not precisely the same as the rules used in BUP since Matsumoto et al. compile their rules a little further to take advantage of the first argument and predicate name indexing used in Prolog.

One could view the strategy taken in this paper as rather similar. Running a tree traversal program in reverse is like randomly generating phrase structure. Then these randomly generated structures are tested against the constraints, i.e., the phrase structure rules. What I have shown here, is that the decision as to where to fold in the constraints is very significant. Folding in the constraints at different positions actually gives completely different parsing strategies.